# Optimizing News Text Classification with Bi-LSTM and Attention Mechanism for Efficient Data Processing

Bingyao Liu
University of California, Irvine
Irvine, USA

Jiajing Chen
New York University
New York, USA

Rui Wang
Carnegie Mellon University
Pittsburgh, USA

Junming Huang
Carnegie Mellon University
Pittsburgh, USA

Yuanshuai Luo
Southwest Jiaotong University
Chengdu, China

Jianjun Wei*
Washington University in St. Louis
St Louis, USA

*Abstract*—The development of Internet technology has led to a rapid increase in news information. Filtering out valuable content from complex information has become an urgent problem that needs to be solved. In view of the shortcomings of traditional manual classification methods that are time-consuming and inefficient, this paper proposes an automatic classification scheme for news texts based on deep learning. This solution achieves efficient classification and management of news texts by introducing advanced machine learning algorithms, especially an optimization model that combines Bi-directional Long Short-Term Memory Network (Bi-LSTM) and Attention Mechanism. Experimental results show that this solution can not only significantly improve the accuracy and timeliness of classification, but also significantly reduce the need for manual intervention. It has important practical significance for improving the information processing capabilities of the news industry and accelerating the speed of information flow. Through comparative analysis of multiple common models, the effectiveness and advancement of the proposed method are proved, laying a solid foundation for future news text classification research.

I. INTRODUCTION

With the rapid development and popularization of Internet technology, news information on the Internet has shown an explosive growth trend, allowing people to easily obtain various news information from the Internet. However, this also brings certain challenges, namely how to extract valuable content from the vast amount of online news. The news industry plays a vital role in economyand is closely related to people's lives. Due to the suddenness and urgency of news, it is also necessary to manage the causes and categories of subsequent news in a detailed manner. Therefore, in order to effectively organize and manage online news information, text classification technology is introduced to effectively organize and manage news text information [1]. This method can divide text into different categories according to its content, thereby avoiding information confusion and helping to extract more valuable content.

In the past, people often used manual text classification[2], which was time-consuming and inefficient. This method is even more difficult when faced with a large amount of text information. However, for news, the use of artificial intelligence technologyfor text classification can improve the accuracy and timeliness of information and save manpower . In recent years, due to technological advances, traditional text classification methods are no longer suitable for today's needs. On the contrary, many emerging technologies, such as machine learning, have begun to be widely used in text classification, thereby improving efficiency and reducing labor costs. In addition, with the advancement of technology, many scholars have also begun to explore deep learning in order to obtain more practical applications [3-6].

In the field of news, there are many news text classification tasks. In terms of text expression methods, there are problems such as large amounts of data, unclear classification, and inability to provide rich emotional semantic expressions for texts [7]. This paper uses data mining technology in deep learning to classify news texts, which is conducive to understanding the development of industry prospects and has important significance in understanding industry classification and news sentiment analysis. Based on the above problems, this paper studies news text classification. The main research work is as follows: realize feature extraction and text classification of news texts, and analyze experimental results. The improved BI-LSTM-Attention [8], bidirectional long short-term memory network and attention mechanism combined model are used to classify news texts, and semantic information is obtained from the context sequence of the text through this method; on this basis, an evaluation mechanism based on Attention (weight) is proposed, and the classification judgment of news is realized through the softmax (classification) function, and it is output. Finally, by comparing with several other classic classification algorithms [9-11], corresponding experiments were designed for the proposed algorithm, and comparative experiments were carried out using the TensorFlow platform. The analysis of the experimental results proved that the constructed model helps to improve the classification effect.

## II. RELATED WORK

Recent advancements in deep learning have significantly contributed to the field of text classification, particularly for large-scale and complex datasets such as news articles. This section discusses the research efforts and methodologies that have influenced the development of the Bi-LSTM-Attention model for efficient news text classification, focusing on core deep learning strategies, optimization techniques, and model enhancements relevant to the proposed approach.

Several studies have laid the foundation for optimizing text classification using deep learning techniques. One major contribution involves the use of graph neural networks for text classification optimization [12]. This work highlights how advanced neural networks can process complex data structures, which is particularly relevant to handling the vast and interconnected nature of news data. The application of graph-based optimization enhances the model's ability to categorize text more effectively by leveraging deeper contextual and relational information. The integration of attention mechanisms in neural networks has become a crucial aspect of improving text classification. Research on deep learning transformers has demonstrated the potential of attention mechanisms in enhancing the performance of models by focusing on the most relevant parts of the input data [13]. In the context of news classification, attention mechanisms allow the model to weigh the importance of different words or phrases in a news article, which helps improve accuracy, particularly in understanding context and nuance. Another important area of research related to news classification is emotional analysis, where deep learning models have been applied to interpret complex language structures and sentiments [14]. This is particularly applicable to sentiment-based news classification, where the emotional tone of the text plays a significant role in categorization. Leveraging emotional analysis techniques allows for more fine-grained classification of news content based on public opinion, tone, or sentiment, which is essential for certain types of news categorization.

Pre-trained models have also played a significant role in improving text classification. Comparative studies on pre-trained models for named entity recognition (NER) underscore the importance of fine-tuning these models for specific language tasks [15]. While NER is a distinct problem, the insights gained from these studies, such as the benefits of pre-trained models for feature extraction, directly inform the development of more accurate and efficient news classification systems. In terms of model optimization, the application of adaptive friction techniques in deep learning optimization strategies has shown promising results in improving the convergence and performance of complex models [16]. Such optimization techniques can be beneficial for training deep learning models that deal with large datasets, such as news texts, by enhancing the learning process and reducing the time required to reach optimal performance. Another contribution comes from research on convolutional neural networks (CNNs) enhanced with higher-order numerical difference methods [17], which demonstrates how numerical optimizations can improve feature extraction in neural models. Hybrid models, which combine the strengths of different architectures, have also emerged as an effective solution for processing sequential data. For example, research into hybrid frameworks combining LSTM with statistical models has demonstrated the value of leveraging multiple methods to enhance the prediction and classification accuracy of models [18]. This approach aligns with the Bi-LSTM-Attention model used in this paper, where the LSTM captures sequential dependencies while the attention mechanism highlights critical aspects of the data, leading to more efficient text classification. Lastly, the use of advanced neural networks for handling dynamic and complex datasets has been explored in various domains[19-20]. The adaptation of these techniques to text classification, especially in dynamic environments like news, shows the potential for handling real-time, evolving data [21]. These methods demonstrate the versatility and robustness of deep learning models in addressing diverse challenges, similar to the dynamic nature of news content.

These papers highlight the growing body of research in deep learning for text classification, particularly with regard to optimization techniques, attention mechanisms, and hybrid models. These studies directly inform and support the development of the Bi-LSTM-Attention model. Unlike traditional models that uniformly apply the attention mechanism across all aspects of the input, our model employs a dynamic attention framework that adapts to the semantic importance of each segment of the text, thereby providing a more granular analysis of news articles.

## III. METHOD

When classifying text, the correlation between contexts has a direct impact on the classification results. The stronger the text correlation, the greater the effect on text classification. In recent years, deep learning network technology has developed rapidly, and LSTM neural networks have made great progress in addressing the shortcomings of long-term information dependence, so they have been widely used in the field of natural language processing. However, LSTM networks only have the characteristics of unidirectional features, so this chapter improves this shortcoming and proposes a text semantic association expression method based on BI-LSTM. The improvement of BILSTM is shown in Figure 1.

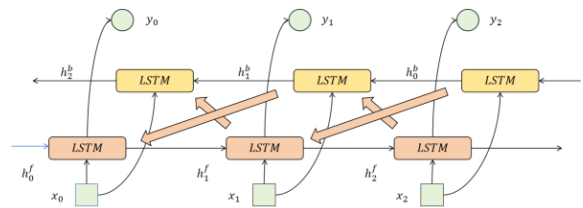

Figure 1 The improvement of BILSTM

Based on this model idea, the BI-LSTM model is improved by combining the contextual semantic feature information of news, so as to construct news text classification method based on BI-LSTM-Attention, and introduce a weight-based classifier, and introduce the weight feature into the BI-LSTM model. The softmax classifier can handle multi-classification problems and complete the text classification of news, so as to better realize the effective classification of news text. The overall framework of the model is shown below

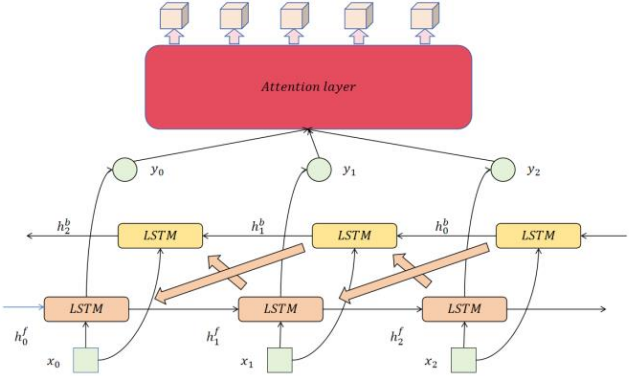

Figure 2 network architecture

By introducing the attention mechanism into the classification of news text, the more valuable feature information in the news text can be effectively preserved, thereby reducing the impact of less relevant or irrelevant feature information. This algorithm can better reflect the key information in the text than conventional algorithms. The news classification model can realize automatic learning of text features, which is very helpful for mining implicit semantic information.

First, the method finds the sequence as the intermediate state, then calculates its Attention probability and assigns it different weights. This algorithm adds the attention mechanism to the BI-LSTM network and uses the median of the LSTM encoder input results to selectively learn the input features. The formula is:

$$u_{it} = \tanh(W_w h_u + b_w)$$
$$\alpha_{it} = \frac{\exp(u_{it}^T u_n)}{\sum_t \exp(u_i^T u_v)} \quad (1)$$
$$s_{it} = \sum_t \alpha \cdot h_H$$

Finally, by using the softmax classifier, we can normalize the feature vector of the text to achieve prediction classification. The formula is expressed as:

$$y = soft\max(W_v v_t + b_v) \quad (2)$$

In order to improve the generalization performance of the model, regularization technology is introduced and the model is improved, which is expressed as:

$$J(\theta) = \frac{1}{m}\sum_{i=1}^{m}(t_i \log(y_i)) + \lambda \|\theta\|_F^2 \quad (3)$$

After the node is updated via the graph attention layer (GAL), the resulting vector output is utilized to derive the neighbor node vectors with varying contribution weights. It is well recognized that graph convolutional neural networks (GCNs) excel in handling graph data rich in dependencies, making them ideal for refining the representation of nodes within a graph. By aggregating the hidden vectors obtained from the GAL through the GCN, the model not only refines the scope of aggregation for aspect terms and sentiment polarity expressions but also streamlines the integration of neighboring nodes based on their differential contributions. This process ensures that the aggregation is more targeted and efficient, allowing for a more nuanced representation of the node's context within the network. The updated aggregation of the node, facilitated by the GCN, is formulated in a way that emphasizes the importance of weighted contributions from its neighbors, thereby enhancing the overall representation and capturing more precise relational information.

$$h_i = \text{Re } LU(\sum_{j \in N_i} A_{ij} W^{(l)} h_i^a + b^l) \quad (4)$$

This paper selects an improved BI-LSTM-Attention model to ensure that the method can extract features while obtaining more feature information. In the specific experiment, the Word2vec model is first used to train the word vector and extract its features. Then the obtained word vector matrix is input into the BI-LSTM-Attention model, and the classification operation is implemented on the data set. Relevant experiments are carried out, and the Adam algorithm and the dropout method are used.

IV. EXPERIMENT

A. Experimental setup

This experiment runs on the open-source learning framework TensorFlow, which can make the design of deep learning modules easier and support various computer languages. It also has high flexibility and uses Python to write programs.

In the training phase of the model, the Adam optimal method is used, the sum of squares error is used as the cost function, and the learning rate (Learning rate) is set to 0.001 according to the size of the data set. After multiple trainings, the optimal parameters are selected. Settings: Adjusted the number of iterations to 10; adjusted the embedded dimension of vocabulary to 200. Adding L2 regular terms can reduce the number of model parameters, remove some output features, and improve the generalization ability of the model. In order to reduce a certain amount of parameters, the dropout strategy is applied to the input layer and BI-LSTM layer. After multiple rounds of experiments, we found that the performance of the model was significantly improved when the dropout rate was adjusted to 0.5.

B. Evaluation indicators

On this basis, we will use accuracy (Precision), recall (Recall) and F1 (F_score) to evaluate the algorithm, so as to improve the credibility and effectiveness of the algorithm.

Accuracy refers to the proportion of correctly classified texts to all texts, and focuses on the overall evaluation effect. Its formula is:

$$precision = \frac{a}{a+b} \times 100\% \quad (5)$$

Recall is a measure that indicates the proportion of correctly identified positive instances out of the total number of actual positives in the dataset. In other terms, it reflects the model's effectiveness in finding all the relevant cases. The recall metric is crucial for understanding how well the model performs in terms of capturing true positive predictions. Its calculation involves dividing the number of true positive predictions by the sum of true positives and false negatives. This formula helps to quantify the model's sensitivity or completeness, ensuring that we can gauge its capability to accurately identify positive samples within the given data.

$$\text{Re}call = \frac{a}{a+c} \times 100\% \quad (6)$$

For the classifier, although 80% of the text belongs to a similar category, if all the text is marked as similar, it will not have any practical application value. To avoid this problem, this paper uses the evaluation index F1 value, which is expressed as:

$$F1 = \frac{2 \times \Pr ecision \times \text{Re}call}{\Pr ecision + \text{Re}call} \quad (7)$$

### C. Experimental Results

According to the different news titles and contents, the classification results are calculated at a ratio of 2:8, and the classification results are divided into four categories according to different categories. Each category is assigned a corresponding score. We can evaluate the reliability of the news source. Through experiments, the training results of the BI-LSTM and BI-LSTM-Attention models are obtained. The loss value changes with the number of iterations. The model effect is judged according to the loss value. The change in loss value is shown in Figure 3.

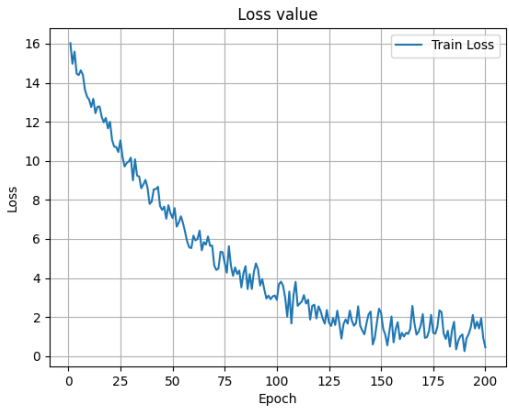

Figure 3(a) BI-LSTM loss function

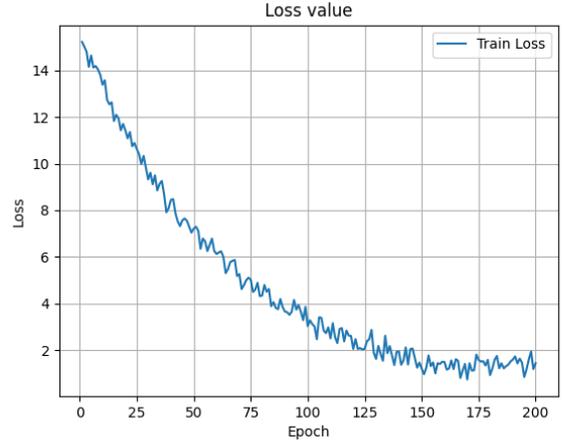

Figure 3(b) BI-LSTM-Attention loss function

This paper designs a series of comparative experiments to thoroughly evaluate the effectiveness of the proposed model, benchmarking it against traditional text classification algorithms prevalent in the field of natural language processing. These include the bidirectional long short-term memory network with attention mechanism (BI-LSTM-Attention), the recurrent neural network (RNN), the convolutional neural network (CNN), the long short-term memory network (LSTM), and the bidirectional long short-term memory network (BI-LSTM). By conducting these experiments on a shared news text dataset, the HuffPost News Dataset [22], the study aims to systematically assess the performance and relative strengths of each model, providing a comprehensive understanding of their effectiveness in real-world applications. The dataset is processed using the Linked Data methodology, which emphasizes the seamless integration of different data formats, a crucial element in academic research [23]. This approach enhances the interconnection of data, enabling more efficient linking and cross-referencing within a unified framework. In domains such as machine learning and artificial intelligence, where the integrity and structure of data are vital, Linked Data provides a robust foundation for organizing information, ultimately leading to more effective model training and improved predictive accuracy. This evaluation not only highlights the proposed model's capabilities but also offers insights into how it compares against established methodologies, making the findings pertinent for both academic research and practical implementation.

Table 1  Model experimental results in the dataset

| Model | Precision | Recall | F1 |
|---|---|---|---|
| RNN | 0.876 | 0.857 | 0.863 |
| CNN | 0.883 | 0.865 | 0.871 |
| LSTM | 0.896 | 0.873 | 0.886 |
| BILSTM | 0.905 | 0.876 | 0.891 |
| Attention | 0.911 | 0.904 | 0.905 |
| Ours | 0.923 | 0.991 | 0.939 |

In our study, we evaluated various models based on precision, recall, and F1 score. Our proposed model outperformed others, achieving a precision of 0.923, recall of 0.991, and an F1 score of 0.939, effectively minimizing false negatives while maintaining high precision. In comparison, RNN recorded precision of 0.876 and recall of 0.857, resulting in an F1 score of 0.845. CNN slightly improved in recall but not in F1 score. LSTM showed enhanced capabilities in processing sequential data with better precision and recall, achieving an F1 score of 0.886. BiLSTM improved precision but had limited impact on recall, with an F1 score of 0.891. Our model demonstrated significant advantages in both precision and recall, leading to the best F1 score, indicating its superior performance in identifying correct positives and controlling false alarms. This suggests promising directions for future research, especially on improving recall while preserving or enhancing precision.

## V. CONCLUSION

To address challenges in news text classification, such as large data volumes, ambiguous categorization, and the lack of expressive emotional semantics, this paper introduces an innovative solution leveraging data mining within deep learning frameworks. The core research centers on the feature extraction and classification of news articles, followed by a comprehensive analysis of experimental outcomes. By employing an enhanced BI-LSTM-Attention model, which integrates a bidirectional long short-term memory network with an attention mechanism, the model effectively captures semantic information from the text's contextual sequences. Furthermore, an attention-based evaluation system is proposed to facilitate accurate news classification. Experimental results demonstrate that this model significantly outperforms traditional approaches, particularly in improving classification accuracy and recall rates, thereby offering robust insights into industry trends and development. This approach holds substantial relevance for both industry classification and sentiment analysis in news content.